\documentclass[runningheads]{llncs}
\usepackage{graphicx}
\usepackage{tikz}
\usepackage{comment}
\usepackage{amsmath,amssymb} % define this before the line numbering.
\usepackage{color}

\begin{document}
% \renewcommand\thelinenumber{\color[rgb]{0.2,0.5,0.8}\normalfont\sffamily\scriptsize\arabic{linenumber}\color[rgb]{0,0,0}}
% \renewcommand\makeLineNumber {\hss\thelinenumber\ \hspace{6mm} \rlap{\hskip\textwidth\ \hspace{6.5mm}\thelinenumber}}
% \linenumbers
\pagestyle{headings}
\mainmatter
\def\ECCVSubNumber{100}  % Insert your submission number here

\title{2nd Place Solution to ECCV 2020 VIPriors Object Detection Challenge} % Replace with your title

% INITIAL SUBMISSION 
%\begin{comment}
%\titlerunning{ECCV-20 submission ID \ECCVSubNumber} 
%\authorrunning{ECCV-20 submission ID \ECCVSubNumber} 
%\author{Anonymous ECCV submission}
%\institute{Paper ID \ECCVSubNumber}
%\end{comment}
%******************

% CAMERA READY SUBMISSION
%\begin{comment}
\titlerunning{2nd Place Solution to VIP Object Detection}
% If the paper title is too long for the running head, you can set
% an abbreviated paper title here

\author{Yinzheng Gu \and Yihan Pan \and Shizhe Chen}
\authorrunning{Gu et al.}
% First names are abbreviated in the running head.
% If there are more than two authors, 'et al.' is used.
%
\institute{Jilian Technology Group, Shanghai, China \\
\email{\{guyinzheng, panyihan, chenshizhe\}@videopls.com}}
%\end{comment}
%******************
\maketitle

\begin{abstract}
In this report, we descibe our approach to the ECCV 2020 VIPriors Object Detection Challenge which took place from March to July in 2020. We show that by using state-of-the-art data augmentation strategies, model designs, and post-processing ensemble methods, it is possible to overcome the difficulty of data shortage and obtain competitive results. Notably, our overall detection system achieves 36.6$\%$ AP on the COCO 2017 validation set using only 10K training images without any pre-training or transfer learning weights ranking us 2nd place in the challenge.

\end{abstract}

\section{Introduction and Challenge Overview}

In order to save training data and reduce energy consumption, the 1st Visual Inductive Priors for Data-Efficient Deep Learning Workshop was introduced as an ECCV 2020 workshop to promote data efficiency. As part of the workshop, four challenges were offered covering various popular research areas of computer vision such as image classification, semantic segmentation, object detection, and action recognition. Each challenge uses a small fraction of the publicly available benchmark dataset with the strict rule that models must be trained from scratch and no external data is allowed. In particular, the common practice of first pre-training the model on a classification dataset such as ImageNet followed by fine-tuning on the task-specific dataset is not applicable. For the object detection track, a subset of the MS COCO 2017 \cite{LMBHPRDZ2014} dataset is used. Specifically, the original COCO dataset contains approximately 118K training images, 5K validation images, and 40K test images split into test-dev and test-challenge. For this challenge, the training and validation sets contain 5,873 and 4,946 images respectively from the original training set, and 4,952 images of the original validation set are used to form the test set, which is what the final ranking is based on. As pre-training is not allowed, our method is based on the following approach.  

Among modern object detectors, DSOD \cite{SLLJCX2017} was the first attempt to successfully train a network from scratch by introducing a set of principles specifically designed for this purpose. Later on, Zhu et al. introduced ScratchDet \cite{ZZWWSBM2019} by redesigning the backbone and intergrating Batch Normalization (BN) \cite{IS2015} in both the backbone and the head of the detector to help convergence when training from scratch. Although satisfactory performance could be obtained, a major drawback of detectors such as DSOD and ScratchDet is the requirement of special designs. For instance, both of them are of one-stage type detectors, and it is usually the case that two-stage detectors achieve better performance in terms of accuracy when speed and memory are not critical. In this direction, the work that is most influential to us is of He et al. \cite{HGD2019} in which it was shown that the performance of training from scratch can be on par with its pre-training couterpart across a variety baseline systems using their hyper-parameters as long as proper normalization techniques such as Group Normalization (GN) \cite{WH2018} or Synchronized BN (SyncBN) \cite{PXLJZJYS2018} are used in place of BN and the model is trained for a sufficiently longer period of time. As the goal of this challenge is for accuracy, we will only be considering two-stage detectors in our experiments.

\section{Method Descriptions and Preliminary Results}

In this section, we provide an overview of our methods and present some preliminary results, full results are given in the next section.

\subsection{Implementation Details}

All of our experiments are carried out using the open source object detection toolbox MMDetection \cite{CWPCXLSFLXZCZCZLLZWDWSOLL2019}. During training, images are resized such that the maximum size is $1333 \times 800$ pixels with 800 being the shorter edge and 1333 being the longer edge without distorting the original aspect ratio. We train the models on 4 or 8 NVIDIA Tesla P100 or V100 GPUs using the so-called $6\times$ schedule from \cite{HGD2019}; that is, 73 epochs in total where the first epoch is used as warmup. The initial learning rate is set to be $0.00125$ $\times$ batch size, where batch size is one of $\{4, 8, 16\}$ depending on model complexity, memory requirement, and GPU type. The learning rate is decreased by a scale of $1/10$ at epoch 65 and 71, respectively, and SGD is used as optimizer with momentum 0.9 and weight decay 0.0001. By default, images are flipped horizontally with probability 0.5.

\subsection{Baseline Results}

For the baseline system, instead of using Faster R-CNN \cite{RHGS2015} and Feature Pyramid Networks (FPN) \cite{LDGHHB2017} with either ResNet \cite{HZRS2016} or ResNeXt \cite{XGDTH2017} as backbone for preliminary analysis like most other works, we go one step further and add modulated deformable convolution with deformable RoI pooling (DCNv2) \cite{ZHLD2019} to the detector and use Cascade R-CNN \cite{CV2018} as the head since these two modules ususally bring improvement. 

In the first set of experiments, we investigate the effect of pre-training and the choice of backbone on the final result. All models are trained using the training set and tested on the validation set of the challenge.

\begin{figure}
	\centering
	\includegraphics[height=5cm]{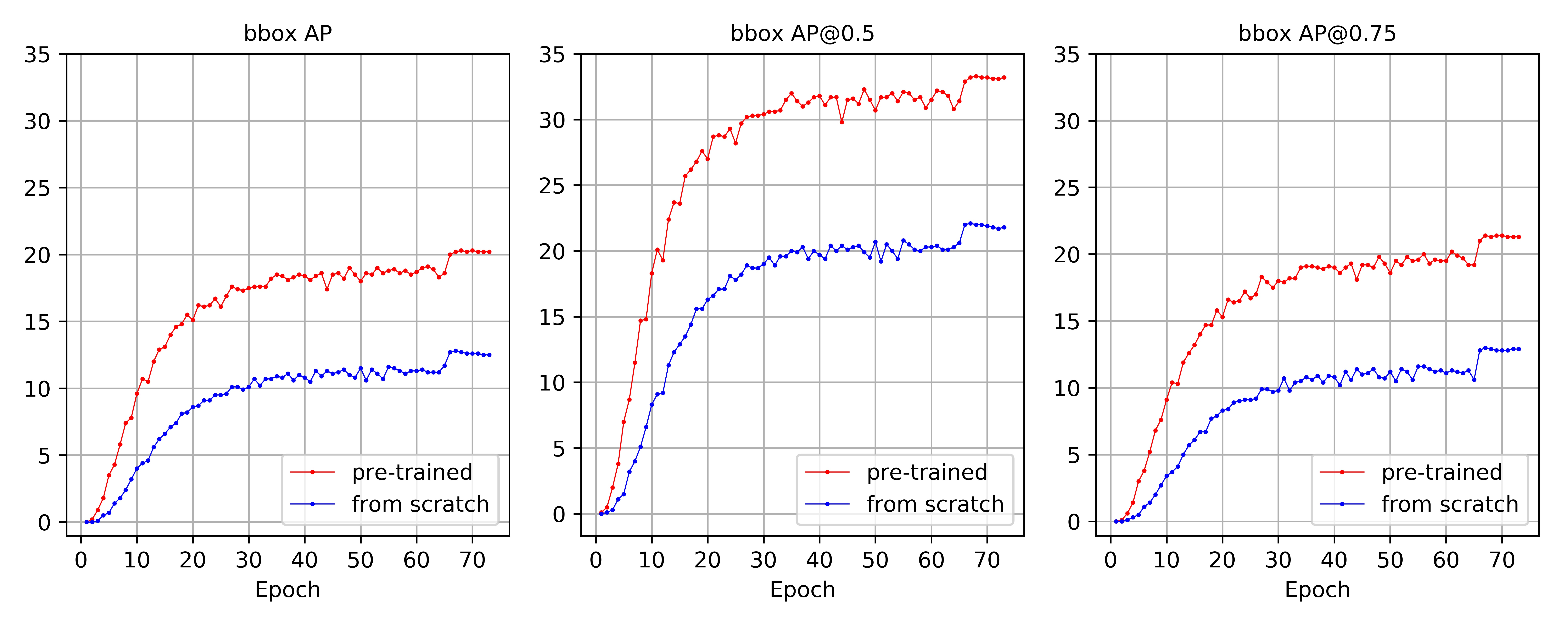}
	\caption{Learning curves of AP (left), AP@0.5 (middle), and AP@0.75 (right) on the COCO2017-VIPriors validation set using Cascade R-CNN with R50-FPN-DCNv2. The results show clear dominance of the model that is pre-trained on ImageNet (20.3 AP) over one that is trained from scratch (12.8 AP). See Table \ref{table:backbone_comparison} for more AP numbers.}
	\label{fig:plot_baseline_pretrained_vs_scratch}
\end{figure}

As can be seen from Figure \ref{fig:plot_baseline_pretrained_vs_scratch}, at this stage pre-training has a significant impact on the performance. Note that although pre-training is not allowed for the challenge, it is still important to perform this experiment for comparison purposes since, according to the discussion section in \cite{HGD2019}, when the training data is reduced to less than 10K images instead of the full COCO dataset, the conclusion that models trained from scratch can perform no worse than those with pre-trained weights no long holds.

\begin{table}
	\begin{center}
		\caption{Results on COCO2017-VIPriors validation set for different backbones.}
		\label{table:backbone_comparison}
		\begin{tabular}{l||ccc||ccc}
			\hline
			Backbone & AP & $\text{AP}_{0.5}$ & $\text{AP}_{0.75}$ & $\text{AP}_{\text{S}}$ & $\text{AP}_{\text{M}}$ & $\text{AP}_{\text{L}}$\\
			\hline
			\hline
			ResNet-50 & \textbf{12.8} & 22.1 & 13.0 & 5.9 & 12.0 & \textbf{17.7} \\
			ResNet-101 & 12.7 & 21.9 & 13.2 & 5.9 & 12.6 & 17.3 \\
			ResNet-152 & \textbf{12.8} & 21.8 & \textbf{13.4} & 5.6 & \textbf{12.8} & 17.6 \\
			ResNeXt-50-32x4d  & 12.3 & 21.5 & 12.7 & 6.0 & 11.6 & 17.1 \\
			ResNeXt-101-32x4d & \textbf{12.8} & 22.1 & 13.1 & \textbf{6.2} & 12.3 & 17.5 \\
			ResNeXt-101-32x8d & 12.6 & 21.7 & 13.0 & 5.9 & 12.2 & 17.1 \\
			ResNeXt-101-64x4d & 12.7 & \textbf{22.2} & 13.0 & 6.0 & 12.3 & 17.1 \\
			\hline
			\hline
			ResNet-50 (pre-trained) & \textbf{20.3} & \textbf{33.3} & 21.3 & \textbf{8.5} & \textbf{20.5} & \textbf{28.8} \\
			ResNeXt-101-64x4d (pre-trained) & 20.2 & 32.6 & \textbf{21.4} & 8.2 & 20.4 & 28.0 \\
			\hline
		\end{tabular}
	\end{center}
\end{table}

In Table \ref{table:backbone_comparison}, we show that at this stage when there is not enough training data, a bigger and more complex backbone such as ResNeXt-101-64x4d has no advantage (in fact a slight disadvantage) over a much smaller network such as ResNet-50. For this purpose, we shall carry out the next set of experiments using ResNet-50 as it trains much faster.

Next, we take advantage of the convenient modular design nature of the MMDetection toolbox and experiment with some of the most recently proposed modules and features added on top of the baseline system. Specifically, we have chosen the IoU-balanced sampling, balanced feature pyramid, and balanced L1 loss components from the Libra R-CNN \cite{PCSFOL2019} framework; the anchor generation and feature adaption components from the Guided Anchoring \cite{WCYLL2019} framework; and the spatial attention mechanisms from the Generalized Attention \cite{ZCZLD2019} framework as enhancements of the current system. Results are summarized in Table \ref{table:detector_comparison} below.

\begin{table}
	\begin{center}
		\caption{Results on COCO2017-VIPriors validation set by adding different modules.}
		\label{table:detector_comparison}
		\begin{tabular}{l||ccc||ccc}
			\hline
			Method & AP & $\text{AP}_{0.5}$ & $\text{AP}_{0.75}$ & $\text{AP}_{\text{S}}$ & $\text{AP}_{\text{M}}$ & $\text{AP}_{\text{L}}$\\
			\hline
			\hline
			Cascade R-CNN with R50-FPN-DCNv2 & 12.8 & 22.1 & 13.0 & 5.9 & 12.0 & 17.7 \\
			+ Libra R-CNN Modules & 15.0 & 24.5 & 15.8 & \textbf{7.5} & 15.0 & 20.6 \\
			+ Guided Anchoring & 15.0 & 24.5 & 15.8 & 7.3 & 14.7 & 21.2 \\
			+ Generalized Attention & \textbf{15.2} & \textbf{24.9} & \textbf{16.0} & 6.7 & \textbf{15.0} & \textbf{22.1} \\
			\hline
		\end{tabular}
	\end{center}
\end{table}

In addition to the above modifications, we also apply some common strategies to further boost performance. It is well-known that data augmentation is an effective way to improve generalization abilities. For this challenge, we make use of the open source Albumentations \cite{BIKPDK2020} library and AutoAugment \cite{ZCGLSL2019} learned policies. For Albumentations, the \verb+ShiftScaleRotate+, \verb+RandomBrightnessContrast+, \verb+RGBShift+, \verb+HueSaturationValue+, \verb+JpegCompression+, \verb+ChannelShuffle+, \verb+Blur+, and \verb+MedianBlur+ operations are used for data pre-processing, whereas for AutoAugment, we directly use the learned \verb+Sub-policy 1+ to \verb+Sub-policy 5+ from \cite[Table 7]{ZCGLSL2019} with equal probability. Furthermore, both multi-scale training and multi-scale testing are included. For training, the size of image is extended from (1333, 800) to the scale range from (1333, 480) to (1333, 960), whereas during inference, images are resized over scales $\{480, 640, 800, 960, 1120, 1280\}$ for the shorter edge and the results are merged together using Soft-NMS \cite{BSCD2017}. Table \ref{table:train_and_test_comparison} below records the overall improvement.

\begin{table}
	\begin{center}
		\caption{Results on COCO2017-VIPriors validation set with data augmentations and multi-scales.}
		\label{table:train_and_test_comparison}
		\begin{tabular}{l||ccc||ccc}
			\hline
			Method & AP & $\text{AP}_{0.5}$ & $\text{AP}_{0.75}$ & $\text{AP}_{\text{S}}$ & $\text{AP}_{\text{M}}$ & $\text{AP}_{\text{L}}$\\
			\hline
			\hline
			Baseline + Libra, GA, Attn & 15.2 & 24.9 & 16.0 & 6.7 & 15.0 & 22.1 \\
			+ Multi-Scale Train & 17.8 & 29.0 & 18.8 & 9.4 & 17.9 & 24.0 \\
			+ Albumentations & 18.2 & 29.8 & 19.2 & 9.2 & 19.0 & 24.1 \\
			+ AutoAugment & 19.3 & 31.5 & 20.6 & 10.2 & 20.1 & 25.9 \\
			+ Multi-Scale Test & \textbf{20.6} & \textbf{32.9} & \textbf{22.0} & \textbf{11.6} & \textbf{21.3} & \textbf{26.8} \\
			\hline
		\end{tabular}
	\end{center}
\end{table}

Finally, using all of the successful methods and strategies above, we experiment on the test set of the challenge for different backbones. In Table \ref{table:baseline_final_comparison} below, we also include results obtained by models learned on the union of training and validation sets, which is allowed by the rules. Note that (1) when only using the training set, the performance is essentially the same on the validation and test sets showing that these two sets have very similar distributions; and (2) when using the union set, bigger models start to show advantage over smaller ones, especially ResNet-152.

\begin{table}
	\begin{center}
		\caption{Results on COCO2017-VIPriors test set using all of the mentioned methods and strategies.}
		\label{table:baseline_final_comparison}
		\begin{tabular}{l|c||ccc||ccc}
			\hline
			Backbone & Data & AP & $\text{AP}_{0.5}$ & $\text{AP}_{0.75}$ & $\text{AP}_{\text{S}}$ & $\text{AP}_{\text{M}}$ & $\text{AP}_{\text{L}}$\\
			\hline
			\hline
			ResNet-50 & train & 20.7 & 32.9 & 22.1 & 10.9 & 21.7 & 27.8 \\
			ResNet-101 & train & 21.0 & 33.0 & 22.3 & 11.7 & 21.8 & 27.3 \\
			ResNeXt-101-32x4d & train & 20.5 & 32.3 & 21.9 & 11.3 & 21.0 & 27.8 \\
			\hline
			\hline
			ResNet-50 & train+val & 27.0 & 41.3 & 28.9 & 15.1 & 28.0 & 35.3 \\
			ResNet-101 & train+val & 27.6 & 41.7 & 29.8 & 15.4 & 28.7 & 36.2 \\
			ResNet-152 & train+val & 28.4 & 42.8 & 30.5 & 15.7 & 29.7 & 37.3 \\
			ResNeXt-101-32x4d & train+val & 27.1 & 41.3 & 29.2 & 15.4 & 28.3 & 36.1 \\
			ResNeXt-101-64x4d & train+val & 27.5 & 42.2 & 29.4 & 15.7 & 28.7 & 36.7 \\
			5-Model Ensemble & train+val & \textbf{28.9} & \textbf{43.7} & \textbf{31.1} & \textbf{16.7} & \textbf{30.2} & \textbf{38.3} \\
			\hline
		\end{tabular}
	\end{center}
\end{table}

\section{Final Results}

For simplicity, we shall denote the detection system from the previous section (including all data augmentation strategies, modules, and multi-scale training and testing) by ``Enhanced Baseline'' (EB). For our final results, we add two more network modifications and data augmentation strategies, which came to our attention during the last month of the challlenge that we found to be very effective.

\begin{figure}
	\centering
	\includegraphics[height=5cm]{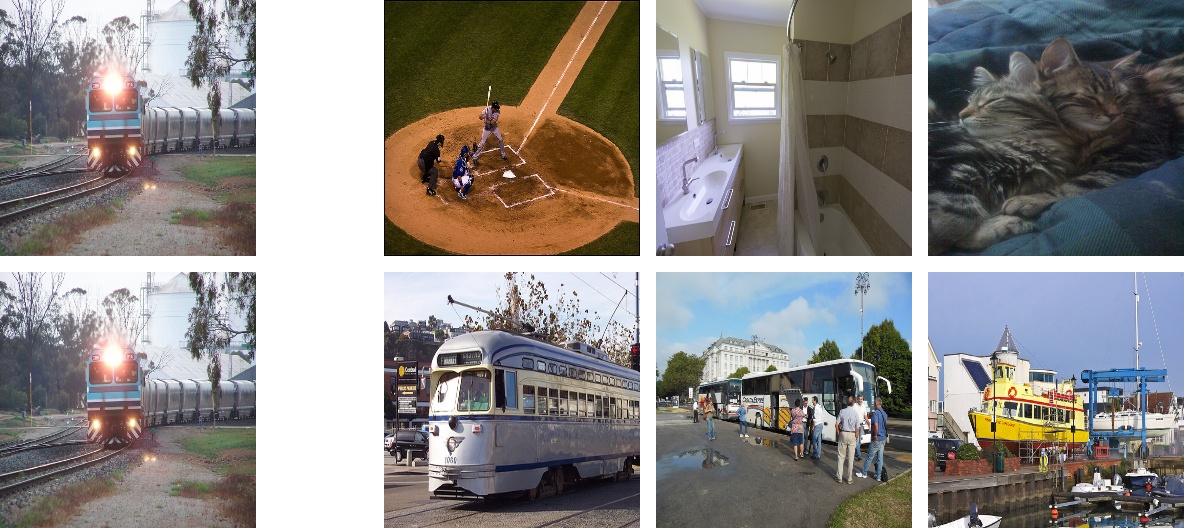}
	\caption{An example demonstrating the effect supercategory-aware selection. In the top, four images are randomly chosen to form either a stitcher or mosaic sample. In the bottom, the leftmost image is used as the query and three other images are selected based on the supercategory of the category of the object in the query image. In this case, we have ``train'' being the category with ``vehicle'' being the supercategory.}
	\label{fig:supercategory_aware_selection_comparison}
\end{figure}

\begin{itemize}
	\item \textbf{TSD} \cite{SLW2020}: This is a recently propoed method aimed at solving the spatial misalignment problem between the classification and regression functions by decoupling the sibling head into task-specific branches via a new operation called task-aware spatial disentanglement (TSD).
	
	\item \textbf{ResNet-D} \cite{HZZZXL2019}: This is a slight model tweak of the ResNet architecture proposed as one of the tricks in \cite{HZZZXL2019} to improve classification accuracy, which turns out to be also beneficial for the task of object detection.
	
	\item \textbf{Stitchers} \cite{CZLLZMXSJ2020}: This is an effective data augmentation strategy with nearly no additional computational cost, where for each sample in a mini-batch, instead of feeding one image like usual four images are resized into smaller sub-images and stitched together and used as a single image.
	
	\item \textbf{Mosaics} \cite{BWL2020}: This method is proposed as part of the YOLOv4 pipeline, which works very similar to that of stitchers that four sub-images make up one image to be used as a sample, except that each sub-image is cropped by a random size selected from a range of scales. In this challenge, we incoporate one more trick that uses the supercategory information as prior knowledge. More specifically, since the 80 categories of the COCO dataset are divided into 12 disjoint supercategories, during the construction of mosaic images, instead of choosing four images at random, one image is first picked at random then depending on the categories of objects appearing in this image, the other three images are chosen whose objects belong to the same supercategory as the categories of the first image. The reason for doing this is that we suspect a model is more likely to be confused by objects from the same supercategory than otherwise. This tweak is denoted by ``Mosaics-SC''. See Figure \ref{fig:supercategory_aware_selection_comparison} for a simple demonstration.
\end{itemize}

In Tables \ref{table:version_d_comparison} and \ref{table:more_data_aug_comparison} below, we verify the effects of the newly added features. On the model side, ResNets with version-D variation consistently outperform plain ResNets while holding all other factors constant for all the experiments we performed. On the data side, the mosaics strategy with supercategory-aware selection appears to be the best.

\begin{table}
	\begin{center}
		\caption{Comparison of a plain ResNet with its version-D counterpart on COCO2017-VIPriors test set. TSD is used for this experiment.}
		\label{table:version_d_comparison}
		\begin{tabular}{l||ccc||ccc}
			\hline
			Backbone & AP & $\text{AP}_{0.5}$ & $\text{AP}_{0.75}$ & $\text{AP}_{\text{S}}$ & $\text{AP}_{\text{M}}$ & $\text{AP}_{\text{L}}$\\
			\hline
			\hline
			ResNet-152 & 27.8 & 42.4 & 29.8 & 16.6 & 28.8 & 36.0 \\
			ResNet-152-D & \textbf{28.8} & \textbf{43.6} & \textbf{31.2} & \textbf{17.1} & \textbf{30.1} & \textbf{36.3} \\
			\hline
		\end{tabular}
	\end{center}
\end{table}

\begin{table}
	\begin{center}
		\caption{Results on COCO2017-VIPriors test set with more data augmentation strategies. EB with ResNet-50 is used for this experiment.}
		\label{table:more_data_aug_comparison}
		\begin{tabular}{l||ccc||ccc}
			\hline
			Data & AP & $\text{AP}_{0.5}$ & $\text{AP}_{0.75}$ & $\text{AP}_{\text{S}}$ & $\text{AP}_{\text{M}}$ & $\text{AP}_{\text{L}}$\\
			\hline
			\hline
			train+val & 27.0 & 41.3 & 28.9 & 15.1 & 28.0 & 35.3 \\
			train+val w/ stitchers & 28.8 & 43.4 & 30.8 & 17.0 & 30.4 & 37.3 \\
			train+val w/ mosaics & 29.6 & 44.5 & 31.8 & 17.2 & 31.0 & 38.7 \\
			train+val w/ mosaics-sc & \textbf{30.1} & \textbf{45.2} & \textbf{32.4} & \textbf{17.6} & \textbf{31.2} & \textbf{39.1} \\
			\hline
		\end{tabular}
	\end{center}
\end{table}

Finally, using all of the methods and strategies described above, our final results are given in Table \ref{table:final_comparison}. For model ensembling, we use the recently proposed Top-k Voting NMS (TkV) \cite{GCDMWLZFH2019}, which is found to better (by about $1\%$ to $2\%$ in terms of AP) than Soft-NMS and the traditional NMS.

\begin{table}
	\begin{center}
		\caption{Final results on COCO2017-VIPriors test set. Here M.R. in the third column stands for multiple runs meaning the model is trained several times by varying the random seed, probability for augmentations, or image size, and the results are combined together and treated as a whole. Two $\checkmark$'s in M.R. means, in addition to multiple runs, each run is also trained for a longer period of time that is approximately equal to the $9\times$ schedule according to \cite{HGD2019}.}
		\label{table:final_comparison}
		\begin{tabular}{ccccccccc}
			\hline
			Method $\&$ Backbone & Data Augmentations & M.R. & AP & $\text{AP}_{0.5}$ & $\text{AP}_{0.75}$ & $\text{AP}_{\text{S}}$ & $\text{AP}_{\text{M}}$ & $\text{AP}_{\text{L}}$\\
			\hline
			\hline
			EB + ResNet-50 & Mosaics-SC & $\checkmark$ & 31.1 & 46.4 & 33.7 & 18.1 & 32.3 & 40.5 \\
			EB + ResNet-101 & Mosaics-SC & $\checkmark$ & 31.2 & 46.2 & 33.6 & 18.2 & 33.0 & 40.3 \\
			EB + ResNet-152 & Mosaics-SC & -- & 31.3 & 46.2 & 34.0 & 17.9 & 32.6 & 40.6 \\
			TSD + ResNet-152-D & Mosaics-SC & $\checkmark\checkmark$ & 33.5 & 49.1 & 36.5 & 21.4 & 35.2 & 41.5 \\
			\multicolumn{2}{c}{Ensemble of the above four using TkV} & -- & 34.7 & 50.5 & 37.8 & 21.3 & 36.4 & 43.4 \\
			EB + ResNet-50-D & Stitchers + Mosaics & $\checkmark$ & 33.5 & 50.0 & 36.1 & 21.1 & 35.3 & 42.5 \\
			EB + ResNet-50-D & Stitchers + Mosaics & $\checkmark\checkmark$ & 35.4 & 51.9 & 38.0 & 21.2 & 37.6 & 44.7 \\
			\multicolumn{2}{c}{Ensemble of the above three using TkV} & -- & 36.6 & 52.9 & 40.0 & 22.4 & 38.7 & 45.6 \\
			\hline
			\hline
			\multicolumn{3}{c}{1st Place Team} & 39.4 & 56.3 & 42.7 & 16.2 & 43.5 & 61.1 \\
			\multicolumn{3}{c}{2nd Place Team (ours)} & 36.6 & 52.9 & 40.0 & 22.4 & 38.7 & 45.6 \\
			\multicolumn{3}{c}{3rd Place Team} & 35.1 & 53.0 & 39.2 & 25.0 & 39.3 & 41.5 \\
			\hline
		\end{tabular}
	\end{center}
\end{table}

\section{Conclusions}

In this report, we presented a detection system that utilizes heavy data augmentations, effective modules, and ensemble strategies which allow us to obtain a satisfactory result of 36.6$\%$ on the MS COCO 2017 validation set using very limited training data and no pre-trained weights achieving 2nd place in the ECCV 2020 VIPriors Object Detection Challenge.

% ---- Bibliography ----
%
% BibTeX users should specify bibliography style 'splncs04'.
% References will then be sorted and formatted in the correct style.
%

\bibliographystyle{splncs04}
\bibliography{ms}

\begin{thebibliography}{10}
\providecommand{\url}[1]{\texttt{#1}}
\providecommand{\urlprefix}{URL }
\providecommand{\doi}[1]{https://doi.org/#1}

\bibitem{BWL2020}
Bochkovskiy, A., Wang, C.Y., Liao, H.Y.: Yolov4: Optimal speed and accuracy of
  object detection. arXiv preprint arXiv:2004.10934  (2020)

\bibitem{BSCD2017}
Bodla, N., Singh, B., Chellappa, R., Davis, L.: Improving object detection with
  one line of code. In: ICCV  (2017)

\bibitem{BIKPDK2020}
Buslaev, A., Iglovikov, V., Khvedchenya, E., Parinov, A., Druzhinin, M.,
  Kalinin, A.: Albumentations: Fast and flexible image augmentations. In:
  Information  (2020)

\bibitem{CV2018}
Cai, Z., Vasconcelos, N.: Cascade r-cnn: Delving into high quality object
  detection. In: CVPR  (2018)

\bibitem{CWPCXLSFLXZCZCZLLZWDWSOLL2019}
Chen, K., Wang, J., Pang, J., Cao, Y., Xiong, Y., Li, X., Sun, S., Feng, W.,
  Liu, Z., Xu, J., Zhang, Z., Cheng, D., Zhu, C., Cheng, T., Zhao, Q., Li, B.,
  Lu, X., Zhu, R., Wu, Y., Dai, J., Wang, J., Shi, J., Ouyang, W., Loy, C.,
  Lin, D.: Mmdetection: Open mmlab detection toolbox and benchmark. arXiv
  preprint arXiv:1906.07155  (2019)

\bibitem{CZLLZMXSJ2020}
Chen, Y., Zhang, P., Li, Z., Li, Y., Zhang, X., Meng, G., Xiang, S., Sun, J.,
  Jia, J.: Stitcher: Feedback-driven data provider for object detection. arXiv
  preprint arXiv:2004.12432  (2020)

\bibitem{GCDMWLZFH2019}
Guo, R., Cui, C., Du, Y., Meng, X., Wang, X., an~J.~Zhu, J.L., Feng, Y., Han,
  S.: 2nd place solution in google ai open images object detection track 2019.
  arXiv preprint arXiv:1911.07171  (2019)

\bibitem{HGD2019}
He, K., Girshick, R., Doll\'{a}r, P.: Rethinking imagenet pre-training. In:
  ICCV  (2019)

\bibitem{HZRS2016}
He, K., Zhang, X., Ren, S., Sun, J.: Deep residual learning for image
  recognition. In: CVPR  (2016)

\bibitem{HZZZXL2019}
He, T., Zhang, Z., Zhang, H., Zhang, Z., Xie, J., Li, M.: Bag of tricks for
  image classification with convolutional neural networks. In: CVPR  (2019)

\bibitem{IS2015}
Ioffe, S., Szegedy, C.: Batch normalization: Accelerating deep network training
  by reducing internal covariate shift. In: ICML  (2015)

\bibitem{LDGHHB2017}
Lin, T.Y., Doll\'{a}r, P., Girshick, R., He, K., Hariharan, B., Belongie, S.:
  Feature pyramid networks for object detection. In: CVPR  (2017)

\bibitem{LMBHPRDZ2014}
Lin, T.Y., Maire, M., Belongie, S., Hays, J., Perona, P., Ramanan, D.,
  Doll\'{a}r, P., Zitnick, C.: Microsoft coco: Common objects in context. In:
  ECCV  (2014)

\bibitem{PCSFOL2019}
Pang, J., Chen, K., Shi, J., Feng, H., Ouyang, W., Lin, D.: Libra r-cnn:
  Towards balanced learning for object detection. In: CVPR  (2019)

\bibitem{PXLJZJYS2018}
Peng, C., Xiao, T., Li, Z., Jiang, Y., Zhang, X., Jia, K., Yu, G., Sun, J.:
  Megdet: A large mini-batch object detector. In: CVPR  (2018)

\bibitem{RHGS2015}
Ren, S., He, K., Girshick, R., Sun, J.: Faster r-cnn: Towards real-time object
  detection with region proposal networks. In: NIPS  (2015)

\bibitem{SLLJCX2017}
Shen, Z., Liu, Z., Jiang, Y.G., Chen, Y., Xue, X.: Dsod: Learning deeply
  supervised object detectors from scratch. In: ICCV  (2017)

\bibitem{SLW2020}
Song, G., Liu, Y., Wang, X.: Revisiting the sibling head in object detector.
  In: CVPR  (2020)

\bibitem{WCYLL2019}
Wang, J., Chen, K., Yang, S., Loy, C., Lin, D.: Region proposal by guided
  anchoring. In: CVPR  (2019)

\bibitem{WH2018}
Wu, Y., He, K.: Group normalization. In: ECCV  (2018)

\bibitem{XGDTH2017}
Xie, S., Girshick, R., Doll\'{a}r, P., Tu, Z., He, K.: Aggregated residual
  transformations for deep neural networks. In: CVPR  (2017)

\bibitem{ZCZLD2019}
Zhu, X., Cheng, D., Zhang, Z., Lin, S., Dai, J.: An empirical study of spatial
  attention mechanisms in deep networks. In: ICCV  (2019)

\bibitem{ZHLD2019}
Zhu, X., Hu, H., Lin, S., Dai, J.: Deformable convnets v2: More deformable,
  better results. In: CVPR  (2019)

\bibitem{ZZWWSBM2019}
Zhu, X., Zhang, S., Wang, X., Wen, L., Shi, H., Bo, L., Mei, T.: Scratchdet:
  Training single-shot object detectors from scratch. In: CVPR  (2019)

\bibitem{ZCGLSL2019}
Zoph, B., Cubuk, E., Ghiasi, G., Lin, T.Y., Shlens, J., Le, Q.: Learning data
  augmentation strategies for object detection. arXiv preprint arXiv:1906.11172
   (2019)

\end{thebibliography}
\end{document}